\documentclass[11pt,a4paper]{article}
\usepackage[hyperref]{acl2019}
\usepackage{times}
\usepackage{latexsym}

\usepackage{tabularx}

\aclfinalcopy 

\usepackage{hyperref}       
\usepackage{url}            
\usepackage{booktabs}       
\usepackage{amsfonts}       
\usepackage{nicefrac}       
\usepackage{microtype}      
\usepackage{natbib}
\usepackage{booktabs}       
\usepackage{balance}
\usepackage{multirow}
\usepackage{graphicx}
\usepackage{subcaption}
\usepackage{amsmath}
\usepackage{float}

\newcommand{\ie}{{\em i.e.}, }
\newcommand{\eg}{{\em e.g.}, }
\newcommand{\noi}{\noindent}
\newcommand{\bi}{\begin{itemize}}
	\newcommand{\ei}{\end{itemize}}

\title{Sarcasm Detection: A Comparative Study}
\author{Hamed Yaghoobian\hspace{1em} Hamid R. Arabnia\hspace{1em} Khaled Rasheed\\ 
        Department of Computer Science\\
        University of Georgia\\
        Athens, GA, 30602, USA \\
        \texttt{\{hy,hra,khaled\}@uga.edu}\\
  }

\date{}
\begin{document}
	
\maketitle
\begin{abstract}
	\noi Sarcasm detection is the task of identifying irony\footnote{Irony is considered an umbrella term that also covers sarcasm; distinguishing between these two rhetoric devices is a further challenge for figurative language processing \citep{farias2016irony}. In short, sarcasm often bears an element of scorn and derision that irony does not \citep{lee1998differential}.} containing utterances in sentiment-bearing text. However, the figurative and creative nature of sarcasm poses a great challenge for affective computing systems performing sentiment analysis. This article compiles and reviews the salient work in the literature of automatic sarcasm detection. Thus far, three main paradigm shifts have occurred in the way researchers have approached this task: 1) semi-supervised pattern extraction to identify implicit sentiment, 2) use of hashtag-based supervision, and 3) incorporation of context beyond target text. In this article, we provide a comprehensive review of the datasets, approaches, trends, and issues in sarcasm and irony detection.  
\end{abstract}

\section{Introduction}
Sarcasm poses a major challenge for sentiment analysis models \citep{liu2010sentiment}, mainly because sarcasm enables one speaker or writer to conceal their true intention of contempt and negativity under a guise of overt positive representation. Thus, recognizing sarcasm and verbal irony is critical for understanding people's actual sentiments and beliefs \citep{maynard2014cares}. The figurativeness and subtlety inherent in its sentiment display, a positive surface with a contemptuous intent (\eg \textit{``He has the best taste in music!''}), or a negative surface with an admiring tone (\eg \textit{``She always makes dry jokes!''}), makes the task of its identification a challenge for both humans and machines. 

Evidently, sarcasm and irony are well-studied phenomena in linguistics, psychology, and cognitive science. In this article, we do not survey the several representations and taxonomies of sarcasm in linguistics \citep{campbell2012there, camp2012sarcasm, ivanko2003context, eisterhold2006reactions, wilson2006pragmatics}, and focus on a descriptive account of the computational attempts at automatic sarcasm detection. Empirical studies of this linguistic device refer to methods to predict if a given user-generated text is sarcastic or not. From a computational perspective, this task is formulated as a \textit{binary classification} problem. Previous research on automated sarcasm detection has primarily focused on lexical, pragmatic resources \citep{kreuz2007lexical} along with interjections, punctuation, sentimental shifts, etc., found in sentences. Nonetheless, sarcasm is often manifested implicitly with no expressed lexical cues. Its identification is reliant on common sense and connotative knowledge that come naturally to most humans but makes machines struggle when extra-textual information is essentially required. Sarcastic utterances are often expressed in such nuanced ways that should be distinguished from a similar phenomenon called \textit{humble-bragging}, which is a self-representational verbal strategy that appears as a complaint concealed within a bragging \citep{wittels2012humblebrag}, as in \textit{``I am a perfectionist at times, it is so hard to deal with''}. To the best of our knowledge, there have been few computational studies that distinguish sarcasm from humble-bragging. 

The remainder of this article is organized as follows. We split the literature along two discernible foci, content- and context-based methods discussed in Sections \ref{content} and \ref{context} respectively, and then classify empirical approaches to sarcasm detection within each section into rule-based, statistical, and deep learning-based.

\section{Content-based methods}
\label{content}
Models investigated in this section base their identification of sarcasm on lexical and pragmatic indicators in English\footnote{Most research in sarcasm detection exists for English. Nonetheless, research in the following languages has been reported also: Utalian, Czech, Dutch, Greek, Indonesian, Chinese, and Hindi.} language use on social media. There is a myriad of novel and intuitive attempts in the literature that fall in this category. We review and categorize studies in this section based on approaches \ref{rule-based} (rule-based, semi-supervised and unsupervised), and features \ref{content-features} (n-gram, sentiment, pragmatics, and patterns) used. 
\subsection{Rule-based}
\label{rule-based}
Rule-based attempts look for evidence and indicators of sarcasm and rely on those in forms of rules. \citet{veale2010detecting} look for sarcastic similes (\eg ``as private as a park-bench'') in the specific query pattern of \textit{``as * as a *''} on Google and using a nine-step approach reveal that 18\% of unique similes are ironical.

Hashtags (or their equivalent, given the social media platform) have been utilized by users to denote sarcasm on Twitter (\eg \#sarcasm, \#not) or on Reddit (\eg /s). Or similarly, if the sentiment of a hashtag does not comply with the rest of the sentence, it is labeled as sarcastic. 

\citet{bharti2015parsing} use a combination of two approaches in their study of sarcasm. They propose a parsing algorithm that looks for sentiment-bearing situations and identifies sarcasm in forms of a contradiction of negative (or positive) sentiment and positive (or negative) situation. They also look for the co-occurrence of interjection hyperbolic words like \textit{``wow''}, \textit{``yay''}, etc. at the start of tweets, and intensifiers like \textit{``absolutely'', ``huge''}  \eg ``\underline{Wow}, that's \underline{a huge discount}, I'm not buying anything!! \#sarcasm.'' Similarly, \citet{riloff2013sarcasm} find a \underline{positive}/\textit{negative} contrast between a sentiment and a situation helpful, and indicative of sarcasm, \eg ``I’m so \underline{pleased} mom \textit{woke me up} with vacuuming my room this morning. :)''. Likewise, \citet{van2018we} speculate that sentiment incongruity within an utterance signifies sarcasm. To this end, they gather all real-world concepts that carry an implicit sentiment and label them with either a ``positive'' or ``negative'' sentiment label. For example, ``going to the dentist'' is often associated with a negative sentiment. Although their model does not surpass the baseline, they highlight the difficulty and importance of incorporating sarcasm detection into sentiment classifiers. They view their efforts as an extension of the seminal work by \citet{greene2009more} to use a concept called \textit{syntactic packaging} to demonstrate the influence of syntactic choices on the perceived implicit sentiment of news headlines. 

One of the earliest work is \citeauthor{tepperman2006yeah}'s that identifies sarcasm in spoken dialogues and relies heavily on cues like laughter, pauses, speaker's gender, and spectral features; their data is restricted to sarcastic utterances that contain the expression `yeah-right'. \citet{carvalho2009clues} improve the accuracy of their sarcasm model by using oral or gestural clues in user comments, such as emoticons, onomatopoeic expressions (\eg \textit{achoo, haha, grr, ahem}) for laughter, heavy punctuation marks, quotation marks, and positive interjections. \citet{davidov2010semi,tsur2010icwsm} utilize syntactic and pattern-based linguistic features to construct their feature vectors.  \citet{barbieri2014modelling} take a similar approach and extend previous work by relying on the inner structure of utterances such as unexpectedness, the intensity of the terms, or imbalance between registers. 

\subsection{Feature sets}
\label{content-features}
In this section, we go over the salient textual features effectively utilized toward the detection of sarcasm. Most studies use bag-of-words to an extent. Nonetheless, in addition to these, the use of several other sets of features have been reported. Table \ref{features} summarizes the main content-based features most commonly used in the literature. We discuss contextual features (\ie features reliant on the codification of information presented beyond text) in Section \ref{context}.

\citet{reyes2012humor} introduce a set of humor-dependent or irony-dependent features related to ambiguity, unexpectedness, and emotional scenario. Ambiguity features cover structural, morphosyntactic, semantic ambiguity, while unexpectedness features gauge semantic relatedness. As we discussed, \citet{riloff2013sarcasm}, in addition to a rule-based classifier, use a set of patterns, specifically positive verbs and negative situation phrases, as features. \citet{liebrecht2013perfect} use bigrams and trigrams and similarly, \citet{reyes2013multidimensional} look into skip-gram and character-level features. In a kindred effort, \citet{ptavcek2014sarcasm} use word-shape and pointedness features. \citet{barbieri2014modelling} include seven sets of features such as maximum/minimum/gap of intensity of adjectives and adverbs, max/min/average number of synonyms and synsets for words in the target text, and so on. \citet{buschmeier2014impact} incorporate ellipsis, hyperbole, and imbalance in their set of features. \citet{joshi2015harnessing} use features corresponding to the linguistic theory of incongruity. The features are classified into two sets: implicit and explicit incongruity-based features.  
\begin{table*}[h!]
\centering
    \small
    \begin{tabularx}{\textwidth}{l X}
    \toprule
    \textbf{Study} & \textbf{Features Used} \\ \midrule
    \citet{reyes2012humor} & Structural, morphosyntactic and semantic ambiguity features\\ \midrule
    \citet{tsur2010icwsm} & Internal syntactic patterns and punctuations \\ \midrule
    \citet{gonzalez2011identifying} & User mentions (replies), emoticons, N-grams, dictionary- and, sentiment-lexicon-based features \\ \midrule
    \citet{liebrecht2013perfect} & N-grams, emotion marks, intensifiers  \\ \midrule
    \citet{hernandez2015applying} & Length of tweet, capitalization, punctuation marks, and emoticons\\ \midrule
    \citet{farias2016irony} & Lexical markers and structural features,  \\ \midrule
    \citet{mishra2016harnessing} & Cognitive features extracted from eye-movement patterns of human readers \\ \midrule
    \citet{joshi2016word} & Features based on word embedding similarity \\ \bottomrule
    \end{tabularx}
    \caption{Features used for Statistical Classifiers}
    \label{features}
\end{table*}

\citet{mishra2016harnessing} propose a novel approach for investigating the salient features of sarcasm in text. They designed a set of gaze-based features such as average fixation duration, regression
count, skip count, etc., based on annotations from their eye-tracking experiments. In addition, they also utilize complex gaze features based on saliency graphs, created by treating words as vertices and saccades (\ie quick jumping of gaze between two positions of rest) between a pair of words as
edges.

\subsection{Learning-based methods}
In the following, we delve more into supervised learning, semi-supervised learning, unsupervised learning, structural and hybrid learning. A brief descriptive account of these approaches toward predictive sarcasm identification in text is given below.

\subsubsection{Supervised learning}
In traditional machine learning approaches, most work on statistical detection of sarcasm has relied on various combinatory forms of Random Forests (RF), Support Vector Machines (SVM), Decision trees (DT), Na{\"\i}ve Bayes (NB) and Neural Networks (NN) \citep{davidov2010semi, joshi2015harnessing, joshi2016word, kreuz2007lexical, reyes2012making, tepperman2006yeah, tsur2010icwsm}. For instance, \citet{gonzalez2011identifying} use SVM with sequential minimal optimization (SMO) and Logistic Regression (LogR), which are usually used toward sentiment analysis, to identify discriminating features. \citet{riloff2013sarcasm} utilize a hybrid SVM system that outperformed the SVM classifier. Similarly, the use of balanced winnow algorithms to determine high-ranking features \cite{liebrecht2013perfect}, Naive Bayes and Decision Trees for multiple pairs of labels among irony, humor, politics, and education \cite{reyes2013multidimensional} and fuzzy clustering for sarcasm detection \cite{mukherjee2017sarcasm} are reported. \citet{bamman2015contextualized} present the use of binary Logistic Regression and SVM-HMM toward incorporating the sequential nature of output labels into a conversation. Likewise, \citet{joshi2015harnessing} report that sequence labeling algorithms are more useful for conversational data as opposed to classification methods. They use SVM-HMM and SEARN as the sequence labeling algorithms. \citet{liu2014sarcasm} present a multi-strategy ensemble learning approach (MSELA) including Bagging, Boosting, etc., to handle the imbalance between sarcastic and non-sarcastic samples.

While rule-based approaches mostly rely upon lexical information and require no training, machine learning invariably makes use of training data and exploits different types of information sources (or features), such as bags of words, syntactic patterns, sentiment information or semantic relatedness. Earliest attempts in this line use similarity between word embeddings as features for sarcasm detection. \citet{ghosh2016fracking} use a combination of convolutional neural networks, LSTM followed by a DNN. \citet{van2018semeval} propose a model that identifies sarcastic tweets and subsequently differentiates the type (out of four classes) of expressed sarcasm. The systems that were submitted for both subtasks represent a variety of neural-network-based approaches (\ie CNNs, RNNs, and (bi-)LSTMs) exploiting word and character embeddings as well as handcrafted features. 

\subsubsection{Semi-supervised learning}
\label{semisupervised}
This form of machine learning, which falls between unsupervised learning and supervised learning, uses a minimal quantity of annotated (labeled) data and a large amount of un-annotated (unlabelled) data during training \cite{tsur2010icwsm}. The presence of the unlabelled datasets and the open access to the unlabelled datasets is the feature that differentiates the semi-supervised from supervised learning. \citet{davidov2010semi} employ a semi-supervised learning approach for automatic sarcasm identification using two different forms of text, tweets from Twitter, and product reviews from Amazon. A total number of 66,000 products and book reviews are collected in their study, and both syntactic and pattern-based features are extracted. The sentiment polarity of 1 to 5 is chosen on the training phase for each training data. The authors report a performance of \%77 precision.

\subsubsection{Unsupervised learning}
\label{unsupervised}
Unsupervised learning in automatic sarcasm identification is still in its infancy, and most approaches are clustering-based, which are mostly applicable to pattern recognition. Nudged by the limitations and difficulties inherent in labeling the datasets (\ie time- and labor-intensivity) in supervised learning methods, researchers seek to eliminate such exertions by focusing on the development of unsupervised models. \citet{nozza2016unsupervised} propose an unsupervised framework for domain-independent irony detection. They build on probabilistic topic models originally defined for sentiment analysis. These models are extensions of the well-known Latent Dirichlet Allocation (LDA) model \cite{blei2003latent}. They propose Topic-Irony model (TIM), which is able to model irony toward different topics in a fully unsupervised setting, enabling each word in a sentence to be generated from the same irony-topic distribution. They enrich their model with a neural language lexicon derived through word embeddings. In a similar attempt, \citet{mukherjee2017sarcasm} utilize both supervised and unsupervised settings. They use Na{\"\i}ve Bayes for supervised and Fuzzy C-means (FCM) clustering for unsupervised learning. Justifiably, FCM does not perform as effectively as NB.

\section{Context-based models}
\label{context}
Making sense of sarcastic expressions is heavily reliant on the background knowledge and contextual dependencies that are formally diverse. As an example, a sarcastic post from Reddit, ``I’m sure Hillary would've done that, lmao.'' requires prior knowledge about the event, \ie familiarly with Hillary Clinton's perceived habitual behavior at the time the post was made. Similarly, sarcastic posts like ``But atheism, yeah *that’s* a religion!'' require background knowledge, precisely due to the nature of topics like \textit{atheism} which is often subject to extensive argumentation and is likely to provoke sarcastic construction and interpretation. The proposed models in this section utilize both content and contextual information required for sarcasm detection. In addition, there has been a growing interest in using neural language models for pre-training for various tasks in natural language processing. We go over the utilization of existing language models \eg BERT, XLNet, etc. toward sarcasm detection in section \ref{lm-context}. 

\citet{wallace2014humans} claim that human annotators consistently rely on contextual information to make judgments regarding sarcastic intent. Accordingly, recent studies attempt to leverage various forms of contextual information mostly external to the utterance, toward more effective sarcasm identification. Intuitively, in the case of Amazon product reviews, knowing the type of books an individual typically likes might inform our judgment: someone who mostly reads and reviews Dostoevsky is statistically being ironic if they write a laudatory review of Twilight. Evidently, many people genuinely enjoy reading Twilight, and so if the review is written subtly, it will likely be difficult to discern the author's intent without this preferential background. Therefore, \citet{mukherjee2017sarcasm} report that including features independent of the text leads to ameliorating the performance of sarcasm models. To this end, studies take three forms of context as feature: 1) author context \citep{hazarika2018cascade,bamman2015contextualized}, 2) conversational context  \citep{wang2015twitter}, and 3) topical context \citep{ghosh2017magnets}. Another popular line of research utilizes various user embedding techniques that encode users' stylometric and personality features to improve their sarcasm detection models \cite{hazarika2018cascade}.  Their model, CASCADE, utilizes user embeddings that encode stylometric and personality features of the users. When used along with content-based feature extractors such as Convolutional Neural Networks (CNNs), a significant boost in the classification performance on a large Reddit corpus is achieved. Similarly to how a user controls the degree of sarcasm in a comment, they extrapolate that the ensuing discourse of comments belonging to a particular discussion forum contains contextual information relevant to the sarcasm identification. They embed topical information that selectively incurs bias towards the degree of sarcasm present in the comments of a discussion. For example, comments on political leaders or sports matches are generally more prone to sarcasm than natural disasters. Contextual information extracted from the discourse of a discussion can also provide background knowledge or cues about the discussion topic. To extract the discourse features, they take a similar approach of document modeling performed for stylometric features. 

\citet{agrawal2020leveraging} formulate the task of sarcasm detection as a sequence classification problem by leveraging the natural shifts in various emotions over the course of a piece of text. \citet{li2020scx} propose a semi-supervised method for contextual sarcasm detection in online discussion forums. They adopt author and topic sarcastic prior preference as context embedding that provides a simple yet representative background knowledge. \citet{nimala2020sentiment} also propose an unsupervised probabilistic relational model to identify common sarcasm topics based on the sentiment distribution of words in tweets.

\subsection{Sarcasm detection using pre-trained language models}
\label{lm-context}
Given the highlighted importance of context to capture figurative language phenomena and the difficulties of data annotation, transfer learning approaches are gaining attention in various domain adaptation problems. In particular, the utilization of pre-trained embeddings such as Global Vectors (GloVe) \cite{pennington2014glove}, and ELMo \cite{peters2018deep} or leveraging Transformer seq2seq methods such as BERT (Bidirectional Encoder Representations from Transformers \cite{devlin2019bert}, RoBERTa \cite{liu2019roberta}, and XLNet \cite{yang2019xlnet}, etc. are witnessing a surge. 

\citet{potamias2020transformer} propose Recurrent CNN RoBERTA (RCNN-RoBERTa), a hybrid neural architecture building on RoBERTA architecture, which is further enhanced with the employment and devise of a recurrent convolutional neural network. They report a performance with an accuracy of \%79 on SARC dataset \cite{khodak2018large}. Similarly, \citet{dadu2020sarcasm} use an ensemble of RoBERTa and ALBERT \cite{lan2019albert} on \textit{Get it \#OffMyChest} dataset \cite{jaidka2020report} achieve a performance of \%85 accuracy with $F1$ score of 0.55. \citet{javdan2020applying} use BERT along with aspect-based sentiment analysis to extract the relation between context dialogue sequence and response. They obtain an F1 score of 0.73 on the Twitter dataset and 0.73 over the Reddit dataset\footnote{Twitter and Reddit datasets used for in this study were provided in the shared task on Sarcasm Detection, organized at Codalab.}. We expect to see more studies geared toward leveraging pre-trained contextual embeddings and transformers toward sarcasm detection in the upcoming years. 

\begin{table}[h!]
\small
\centering
\begin{tabularx}{\columnwidth}{X r r}
\toprule
\textbf{Method} & \textbf{Acc} & \textbf{F1}          \\ \midrule
ELMo \cite{peters2018deep}                                            & 0.70         & 0.70                   \\ \midrule
NBSVM \cite{wang2012baselines}& 0.65         & 0.65                   \\ \midrule
XLnet  \cite{yang2019xlnet}                                          & 0.76         & 0.76                   \\ \midrule
BERT-cased                                             & 0.76         & 0.76                   \\ \midrule
RoBERTa \cite{liu2019roberta}                                         & 0.77         & 0.77                   \\ \midrule
CASCADE \cite{hazarika2018cascade}                   & 0.74         & 0.75                   \\ \midrule
\citet{ilic2018deep}             & 0.79         & \multicolumn{1}{c}{-} \\ \midrule
\citet{khodak2018large}           & 0.77         & \multicolumn{1}{c}{-} \\ \midrule
RCNN-RoBERTa \cite{potamias2020transformer}  & 0.79         & 0.78                   \\ \bottomrule
\end{tabularx}
\caption{State-of-the-art NN classifiers and results on Reddit Politics dataset}
\label{sarc}
\end{table}

\section{Datasets} 
\label{datasets}
This section outlines the datasets used for computational studies on sarcasm detection. Commonly, they are divided into three categories short text (\eg Tweets, Reddits), long text (\eg discussions on forums), transcripts (\eg conversational transcripts of a TV show or a call center). Short text can contain only one (possibly sarcastic) utterance, whereas long text may contain a sarcastic sentence among other non-sarcastic sentences that could potentially function as context. 

\subsection{Short text}
This category of data is the dominant form of expression on social media, mostly as a direct result of restriction on text length. Consequently, this type of text is rife with abbreviations to make efficient use of space on platforms such as Twitter. Two main approaches are utilized toward annotation of tweets: Manual and hashtag-based. \citet{riloff2013sarcasm, maynard2014cares, mishra2016harnessing, ptavcek2014sarcasm} introduce manually annotated datasets of sarcastic utterances. Most annotation approaches in the literature are conducted using hashtags to create labeled datasets. Sarcastic intent in English is commonly and culturally communicated using hashtags such as \#sarcasm, \#sarcastic, \#not. \citet{davidov2010semi, gonzalez2011identifying, reyes2012humor} use hashtag-based datasets of tweets. \citet{liebrecht2013perfect} only uses \#not to collect and label their tweets. While collecting sarcastic tweets using this method is undemanding, the inclusion of non-sarcastic tweets can be challenging since tweets containing \#notsarcastic may not represent a general non-sarcastic text \cite{bamman2015contextualized}. Another approach is to collect the non-sarcastic tweets of users whose sarcastic tweets are also present in the dataset. To ensure collection of true sarcasm, some studies like \citet{fersini2015detecting} manually verified the initial hashtag-based tweets using annotators. 

Reddit is the other popular platform for researchers to collect sarcasm using hashtag ``/s'' (Reddit's equivalent of ``\#sarcasm'' on Twitter). \citet{khodak2018large} present SARC, a large-scale self-annotated corpus for sarcasm that contains more than a million examples of sarcastic/non-sarcastic statements made on Reddit. 

\subsection{Long text}
\citet{lukin2013really} use the Internet Argument Corpus (IAC) \citep{walker2012corpus} which contains a set of 390,704 posts in 11,800 discussions extracted from the online debate site 4forums.com, annotated for several dialogic and argumentative markers, one of them being sarcasm. \citet{reyes2014difficulty} collect a dataset of movie and book reviews, along with news articles marked with sarcasm and sentiment. In an earlier study, \citet{reyes2012making} garner 11,000 reviews of products with sarcastic expressions. \citet{filatova2012irony} present a corpus generation experiment where they collect regular and sarcastic Amazon product reviews. This resulting corpus can be used for identifying sarcasm on two levels: a document and a text utterance, where a text utterance can be as short as a sentence and as long as a whole document. 

\subsection{Transcripts and dialogues}
Sarcasm is often expressed in the context of a conversation, as a response projecting contemptuous intent. \citet{tepperman2006yeah} uses 131 call center transcripts to look for occurrences of “yeah right” as a marker of sarcasm. Similarly, \citet{rakov2013sure} through crowd-sourcing collect sentences from an MTV show called ``Daria.'' \citet{joshi2016word} also present a manually annotated transcript of the popular sitcom ``Friends.'' 

\section{Conclusion}
Sarcasm detection research has seen a significant surge in interest in the past few years, which justifiably calls for an investigation. This article focuses on approaches to automatic sarcasm detection in text. We discern three major paradigms in the history of sarcasm detection research: the use of hashtag-driven supervised learning toward building annotated datasets, semi-supervised pattern extraction to identify implicit sentiment, and the utilization of extra-textual information as context (\eg user's characteristic profiling). While rule-based approaches attempt to capture any indication of sarcasm in the form of rules, statistical methods use features like shifts in sentiment, specific semi-supervised patterns, etc. Deep learning techniques have also been used to incorporate context, \eg additional stylometric features of authors in conversations and the nature of discussion topics. An underlying theme of these past approaches (either in terms of rules or features) is predicated on sarcasm's contemptuous nature. Novel techniques to incorporate contextual insight have also been explored, mostly centered on the emerging direction toward utilizing language models. 
\bibliography{bib}
\bibliographystyle{acl_natbib}
\end{document}